# Application of LSTM architectures for next frame forecasting in Sentinel-1 images time series

Waytehad Rose Moskolaï[a,b] – Wahabou Abdou[a] – Albert Dipanda[a] - Kolyang[b]

[a] Computer Science Department, University of Burgundy, 21078 DIJON Cedex, France,

[b] Computer Science Department, University of Maroua, P.O. Box 46 MAROUA, Cameroon

waytehad-rose_moskolai@etu.u-bourgogne.fr,

{wahabou.abdou, albert.dipanda}@u-bourgogne.fr,

dtaiwe@gmail.com

**RÉSUMÉ.** L'analyse prédictive permet d'estimer les tendances des évènements futurs. De nos jours, les algorithmes Deep Learning permettent de faire de bonnes prédictions. Cependant, pour chaque type de problème donné, il est nécessaire de choisir l'architecture optimale. Dans cet article, les modèles Stack-LSTM, CNN-LSTM et ConvLSTM sont appliqués à une série temporelle d'images radar sentinel-1, le but étant de prédire la prochaine occurrence dans une séquence. Les résultats expérimentaux évalués à l'aide des indicateurs de performance tels que le RMSE et le MAE, le temps de traitement et l'index de similarité SSIM, montrent que chacune des trois architectures peut produire de bons résultats en fonction des paramètres utilisés.

**ABSTRACT.** Predictive analytics allow to estimate future trends of events. Nowadays, Deep Learning algorithms allow making good predictions. However, it is necessary to choose the architecture that produces the most efficient results for each kind of problem. In this paper, the Stack-LSTM, the CNN-LSTM and the ConvLSTM models are applied to a time series of sentinel-1 radar images. The goal is to predict the next occurrence in a sequence of images. Experimental results are evaluated with performance metrics such as the RMSE and MAE loss, the processing time and the SSIM index. The values show that each of the three architectures can produce good results depending on used parameters.

**MOTS-CLÉS :** Apprentissage profond, LSTM, Prédiction, Images satellitaires, Prévision, Changements couverture terrestre.

## 1. Introduction

Predictive analytics are part of data mining that allow estimating future trends of events. It has been significantly developed during the past few years. This scientific domain uses various statistical techniques and particular algorithms that produce predictive models by analyzing past and present information from a time series data-set. Forecasting events and anticipating decision-making is a real necessity in most of activity sectors [1]. Several classical techniques such as auto-regressive or Markov models have been used for a long time in prediction problems and have shown quite satisfactory results However, with the big data advent, it was necessary to design more efficient and complex models in order to consider the huge volume of data and non-linearity aspects in some time series [14]. For this end, several techniques have been developed, including those using artificial neural networks (ANN). These methods have shown very satisfactory results [2] and refer to a computational system based on the functioning of human neurons.

In the literature, several types of Deep Learning architectures are used for time series, and their performance is steadily increasing [4]. For prediction problems involving sequences of data, the Recurrent Neural Networks (RNN) namely the Long Short-Term Memory (LSTM) architectures [12] are generally used for their ability to store the state from previous layers. In models that use basically LSTM architectures with images, predictions are made pixel by pixel and do not take into consideration the spatial distribution of information [3]. However, when the spatiotemporal aspect must be modeled, the Convolutional Neural Networks (CNN) are often used for prediction such as in [9]. To improve accuracy of predictions, researchers in [13] merged functionalities of CNN and LSTM architectures to create a new one, so-called ConvLSTM. Since, several authors have used this new architecture and have obtained better results compared to the use of LSTM or CNN separately [10, 11]. Furthermore, some researchers have instead combined CNN and LSTM networks, used the outputs of one model as input data for the other and obtained good results [15].

Thus, in order to compare performances of three variants of LSTM architectures namely the ConvLSTM, Stack-LSTM and CNN-LSTM in the context of next occurrence prediction in a given time series data, this work proposes to implement each of these architectures. For this purpose, sequences of Sentinel-1 images, representing an area around the Wildlife Reserve of Togodo (WRT) are considered.

This paper is organized as follows. First, in Section 2, Recurrent Neural Networks and LSTM architectures are presented. Then in Section 3, methodology used in this work is described. Finally, in Section 4, the obtained results are presented and discussed before concluding the work in Section 5.

## 2. Recurrent Neural Networks (RNN) and Long Short-Term Memory (LSTM)

Recurrent Neural Networks (RNN) are one of the most advanced supervised Deep Learning architecture and are mainly used with time series. In this architecture, hidden layers are interconnected under time and then they can keep in memory states of previous layers. The recurrent connections add state or memory to the network and allow it to learn and harness the ordered nature of observations within input sequences [5]. However, with long sequences of data, models sometimes faced with the problem of vanishing or exploding gradient. Thus, the network is not able to perform well. To solve the problem of vanishing gradient, a special kind of RNN, the Long Short-Term Memory (LSTM) network has been introduced [12]. Since, much amelioration has been done by researchers and it is now the most popular architecture used for prediction problems.

### 2.1. LSTM Networks

LSTMs networks are a particular variation of RNN. In this architecture, there are four layers that interacting in a special way: information gate, forget gate, input and output gate. Figure 1 shows the configuration of a simple LSTM memory block. Note that in real configurations it may have more gates.

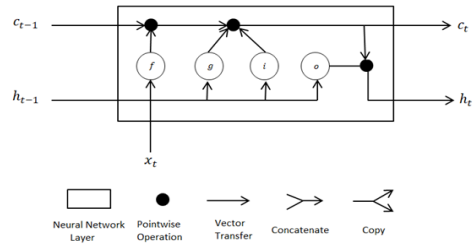

**Figure 1:** *A simple LSTM block memory*

Equations (1) to (6) bellows describe each component in the memory block. Variable descriptions are shown in Table 1. In practice, Vanilla LSTM represents architectures with only one hidden layer while stack-LSTM is made by more hidden layers whose are stacked one on top of another. Sometimes, it can be necessary to allow the LSTM model to learn the input sequence both forward and backwards and concatenate both interpretations. Such model is called a Bidirectional LSTM [14].

$c_t = f_t \odot c_{t-1} + i_t \odot g_t$ (1)  $\quad f_t = \sigma_g(W_f x_t + R_f h_{t-1} + b_f)$ (4)

$h_t = o_t \odot \sigma_c(c_t)$ (2)  $\quad g_t = \sigma_c(W_g x_t + R_g h_{t-1} + b_g)$ (5)

$i_t = \sigma_g(W_i x_t + R_i h_{t-1} + b_i)$ (3)  $\quad o_t = \sigma_g(W_o x_t + R_o h_{t-1} + b_o)$ (6)

## 2.2. ConvLSTM and CNN-LSTM Networks

The use of classical Convolutional Neural Network (CNN) architecture is the best choice when inputs of networks are 2-D or 3-D tensors like images or video [9]. Since LSTMs architectures are more adapted for 1-D Data, new variant of LSTM called Convolutional LSTM or ConvLSTM [13] has been designed. In this architecture, the LSTM cell contains a convolution operation and input dimension of data is kept in output layer, instead of being just a 1-D vector. Matrix multiplication at each gate of classical LSTM is replaced with convolution operation. We can say that ConvLSTM architecture merges capabilities of CNN and LSTM Network. It was normally developed for 2-D spatial-temporal data such as satellite images. However, with some adaptations, it could be used with other types of data. Previous equations (1) to (6) for LSTM are changed for ConvLSTM networks such as in [6], adding convolutional operations in gates.

Another approach for working with spatiotemporal data is to combine CNN and LSTM layers, one block after another. Such architecture is called Convolutional-LSTM (CNN-LSTM) and was originally called Long-term Recurrent Convolutional Network or LRCN model. In the first part of this model, convolutional layers extract important features of input data and results are flattened in 1-D tensor in order to be used as input for the second part of model (LSTM). Finally, before passing data in the last hidden layer, information has to been reshaped in the original form of input data.

| Var. | Definition | Var. | Definition |
|---|---|---|---|
| $c_t$ | Cell state at time t | $x_t$ | Input vector to the LSTM unit |
| $h_t$ | Hidden state at time step t | $W_{i,f,g,o}$ | Input weights for each component |
| $i_t$ | Input gate at time step t | $R_{i,f,g,o}$ | Recurrent weight for each component |
| $f_t$ | Forget gate at time step t | $b_{i,f,g,o}$ | Bias parameters for each component |
| $g_t$ | Cell candidate at time step t | $\sigma_c$ | the gate activation function (by default sigmoid) |
| $o_t$ | Output gate at time step t | $\sigma_g$ | the state activation function (by default tanh) |
| $\odot$ | Hadamart product | | |

**Table 1:** *Definition of variables used in equations (1) to (6)*

## 3. Methodology

### 3.1 Problem formulation

In univariate supervised learning problem, there are always input variables (*X*), output variable (*Y*) and a model which use an algorithm to learn the mapping function from the input to the output *Y=f(X)*. Therefore the goal of predictive models is to estimate the real underlying mapping function and then with new input data (X), models can predict the corresponding output variable (*Y*). Let be $X = \{X_1, X_2, X_3, X_4, \ldots X_{t-1}, X_t, X_{t+1}, X_{t+2}, \ldots, X_{t+n}\}$ a time series of satellites images used for training the models, with shape (*W, H, N*). For a given sequence $Z = \{Z_1, Z_2, \ldots Z_{t-1}, Z\}$ as input for prediction, the objective is to predict *Y=f(Z)*, the output of model such as $Y = Z_{t+1}$. *W, H* and *N* denote respectively the number of rows, column and channels for each image. The objective of this work is thus to implement a sequence-to-one model prediction to forecast the next image occurrence of a given time series. To achieve this, implementation of three variants of LSTM architecture presented in the previous section are done and then, results are compared.

### 3.2 Used Data and preprocessing

Sentinel-1 images covering the wildlife Reserve of Togodo (WRT) in Togo, a West African country are used in this work. The WRT is located between 1°20 and 1°40 East longitudes and between 6°40 and 6°50 North latitude[8]. A set of 158 images in double polarization VV and HV, from September 2016 to May 2019 are downloaded[1]. Before designing the model, some treatments have been done to images as in [8].

Secondly, all images where transformed in gray images, resized to shape *(64, 64, 1)* and pixels value normalized between 0 and 1. Then, about 80% of original dataset was selected as training set and the remaining 20% as test set. Chronological order of images is kept in each data since we are working on forecasting task. In addition, to use a time series for the training of a supervised learning model, data must been first transformed o the form of *(samples, timestep, W, H, features)*. After splitting dataset to correct form by generating *X_train* and *Y_train* arrays as shown in Table 2, a sample equal to (t−timestep) is obtained, corresponding to the batch size for training step. Practically, *timestep* is the number of occurrence in each sample, and here we consider that *timestep = 5*. So, given a time series of five images, the model will forecast the next (sixthly) occurrence. The parameter *features* corresponds to the number of variable to predict. In this work we want to output one image, so this parameter is set to 1.

| X_train | Y_train |
|---|---|
| $[X_1, X_2, X_3, X_4, X_5]$ | $[X_6]$ |
| $[X_2, X_3, X_4, X_5, X_6]$ | $[X_7]$ |
| $[X_3, X_4, X_5, X_6, X_7]$ | $[X_8]$ |
| … | |
| $[X_{t-5}, X_{t-4}, \ldots, X_{t-1}]$ | $[X_t]$ |

**Table 2:** *Overview of training set modelization*

---

[1] www.earth-explorer.usg.org

### 3.3 Structure of models

Before setting parameters for the models, some combinations have been tested in order to have acceptable results which could be improved later. Thus, for ConvLSTM and Stack-LSTM model, only three *ConvLSTM* and three *LSTM* layers have been stacked, associated with *Batchnomalization* and *Dropout* layers to normalize values coming from previous layer and avoid phenomena of over fitting respectively. At the end, we put a *Dense* layer for output. Concerning the CNN-LSTM model, one *2Dconvolutional* layer has been inserted followed by *one maxpooling2D* and one *flatten* layer which allow to obtain 1-D vector to fit in *LSTM* layers. After that, two *LSTM* layers are stacked and finally one *Dense* layer is added for output. Since images are nonlinear objects, the Rectifier Linear Unit *relu* is always used as activator in convolutional layers in order to add linearity in output images. Among the most used optimization functions in literature, *rmsprop, adam*, or *momentum* algorithms are mentioned. Thus, for experimentation step, the *rmsprop* function has been first used since he is usually good choice for RNN. Secondly, the Adaptive Moment Optimization *adam* optimizer which allow to obtain smaller training loss values than *rmsprop* has been used. Indeed, this function combines capabilities of both *rmsprop* and *Momentum* [4]. So *adam* optimizer which is defined by equations (7) to (10), has been definitively adopted for the rest of the work.

$$v_t = \beta_1 * v_{t-1} - (1 - \beta_1) * g_t \quad (7)$$
$$s_t = \beta_2 * s_{t-1} - (1 - \beta_2) * g_t^2 \quad (8)$$
$$\Delta w_t - \eta \frac{v_t}{\sqrt{s_t + \epsilon}} * g_t \quad (9)$$
$$w_{t+1} = w_t + \Delta w_t \quad (10)$$

Where $\eta$ denotes the initial learning rate, $g_t$ gradient at time t, $v_t$ the Exponential Average of Gradient, $s_t$ the Exponential Average of square of gradient, $\beta_1, \beta_2$ are Hyper parameters. Each parameter $w^j$ is replaced by $w$ for more clarity.

The two following functions have been used as loss function: the Root Mean-Square-Error (RMSE) and the Mean Absolute Error (MAE) given by equations (11) and (12)

$$RMSE = \sqrt{\frac{\sum_{i=1}^{n}(P_i - O_i)^2}{n}} \quad (11)$$
$$MAE = \frac{1}{n}\sum_{j=1}^{n}|(P_i - O_i)| \quad (12)$$

Where *n* represents the sample size, $P_i$ represents the predicted values and $o_i$ the observed ones.

## 4. Results and discussion

In this section, performances of three LSTM architectures namely the ConvLSTM, Stack-LSTM and CNN-LSTM are compared for the forecasting task. The study is based on Sentinel-1 image time series representing the wildlife Reserve of Togodo. The RMSE and MAE values described above are considered as performance metrics.

### 4.1 Results

Figure 2, Figure3 and Figure 4 present results of predictions done by the three studied models using MAE as loss function. Figures display the input sequence, the ground truth image and the predicted ones respectively by ConvLSTM, Stack-LSTM andd CNN-LSTM models. For the Figure 2 and Figure 3, the *timstep= 5* while *timestep= 10* in Figure 4.

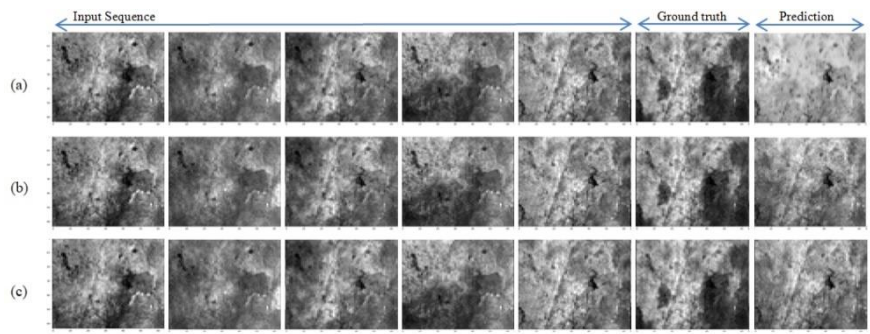

**Figure 2:** *Prediction results based on ConvLSTM, Stack-LSTM and CNN-LSTM (respectively (a), (b), (c)). Timestep = 5, with (64×64) images.*

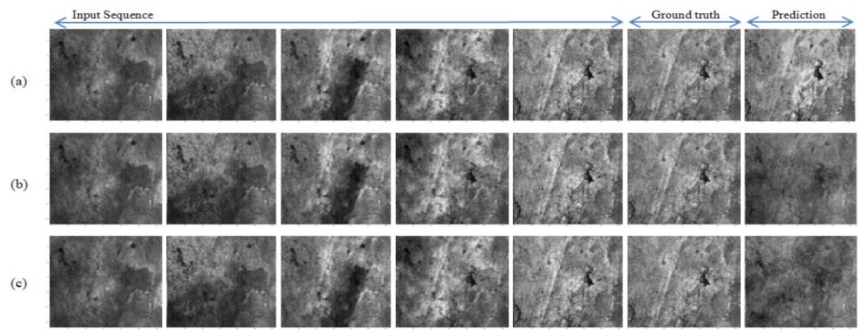

**Figure 3:** *Prediction results based on ConvLSTM, Stack-LSTM and CNN-LSTM (respectively (a), (b), (c)). Timestep = 5, with (128×128) images.*

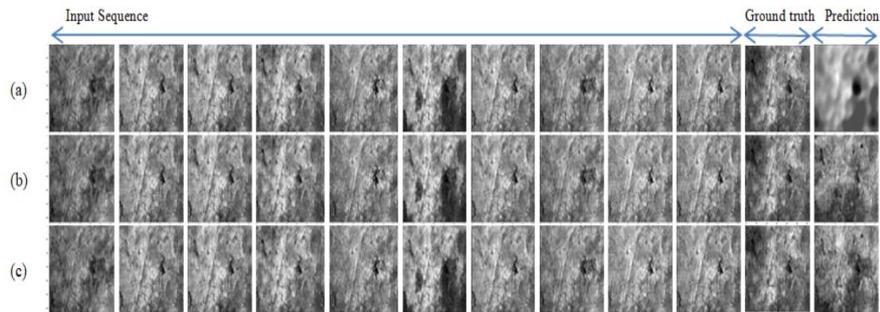

**Figure 4:** *Prediction results based on ConvLSTM, Stack-LSTM and CNN-LSTM (respectively (a), (b), (c)). Timestep = 10, with (64×64) images.*

### 4.2 Discussion

The main goal of machine learning is to produce models which are able to make good predictions on new data. Low training loss values indicate generally that the model is learning well. However, a model with low values does not automatically means that it is efficient. Moreover, values too close to zero indicate sometimes that the model is over-fitting and thus, is not able to perform well on new data. So, it's important to consider others parameters such as validation loss, training time, the Structural Similarity Index SSIM [16] for a better evaluation. SSIM values close to 1 indicate generally good similarities. Equation (13) gives the mathematical formula of this measure which varies between 0 and 1:

$$SSIM(x, y) = \frac{(2\mu_x\mu_y + c_1)(2\sigma_{xy} + c_2)}{(\mu_x^2 + \mu_y^2 + c_1)(\sigma_x^2 + \sigma_y^2 + c_2)} \quad (13)$$

Where $x$ and $y$ are the compared images, $\mu$ represent the means, $\sigma$ represent the variances and $c_1$, $c_2$ are constants.

To study the variation of the output of the different models according to the value of *timestep*, curves representing loss functions are produced by varying *timestep* from 5 to 10. The mean loss and standard deviation are then calculated. With the considered dataset, we can observe that there is no significant change for the training loss when *timestep* values vary, for CNN-LSTM and Stack-LSTM architectures, as represented in Figure 5. However, with the ConvLSTM model, the higher the *timestep*, the higher the training loss values as shown in Figure 5. It is then deduced that the use of ConvLSTM with long sequences is not advisable.

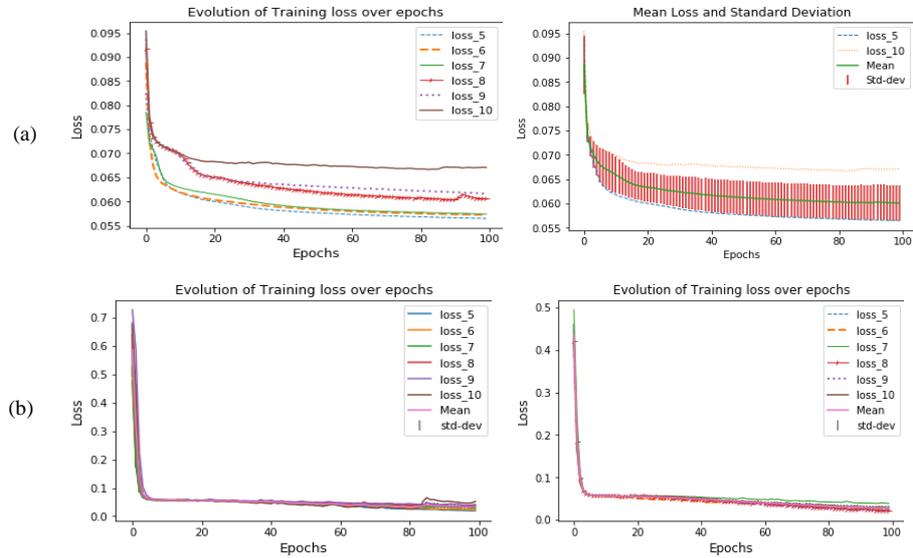

**Figure 5**: *Evolution of training loss (MAE) over epochs depending on timestep (vary from 5 to 10). (a) Left: training loss with ConvLSTM, right: mean and standard deviation. (b) Left: Training loss with Stack-LSTM, right: Training loss with CNN-LSTM.*

Table 3 summarizes different values obtained by the combinations of some parameters used in this study to evaluate models. In order to see the impact of resolution on prediction accuracy, original images were resized to (128×128) before fitting the models. As it is shown in Table 3, evolutions of training and validation loss are different depending on architecture. We can see that the training loss of CNN-LSTM is the lowest with (128×128) images while the Stack-LSTM produces the best training loss value using (64×64) images.

Concerning the training time, we can notice that it increases significantly when the image resolutions are improved for the ConvLSTM model as presented in Table 3. However, the difference is not very big with the Stack-LSTM and CNN-LSTM models. Training time is an important factor to consider when evaluating models and particularly when working on data from remote sensing (earth observation images) which are very large. In addition, quality of output in term of resolution is very important for better interpretation. We can observe in Table 3 that the ConvLSTM architecture takes too much time and consume more memory for training step. With our basic parameters of 100 epochs and *timestep=5*, total training time with this architecture is about four times than time taken by the Stack-LSTM and CNN-LSTM models. Thus, it is deduced that the ConvLSTM is not suitable for complex parameters and large data although SSIM value obtained for (128×128) images using MAE loss is the best comparatively to those obtained with Stack-LSTM and CNN-LSTM.

| Resol. | Architect. | Loss = MAE, timestep = 5 | | | | Loss = RMSE, timestep = 5 | | | |
|---|---|---|---|---|---|---|---|---|---|
| | | Train. Loss | Valid. Loss | Train. Time (s) | SSIM | Train Loss | Valid. Loss | Train. Time (s) | SSIM |
| 128×128 | ConvLSTM | 0,0574 | 0,0952 | 1933 | 0,83 | 0,0752 | 0,0957 | 1904 | 0,78 |
| | Stack-LSTM | 0,0193 | 0,0824 | 304 | 0,67 | 0,0551 | 0,0816 | 401 | 0,69 |
| | CNN-LSTM | 0,0083 | 0,0860 | 315 | 0,60 | 0,0051 | 0,0689 | 402 | 0,78 |
| 64×64 | ConvLSTM | 0,0562 | 0,0975 | 605 | 0,64 | 0,0728 | 0,1205 | 703 | 0,67 |
| | Stack-LSTM | 0,0071 | 0,1161 | 305 | 0,53 | 0,0086 | 0,1235 | 203 | 0,70 |
| | CNN-LSTM | 0,0152 | 0,0847 | 301 | 0,72 | 0,0127 | 0,1286 | 303 | 0,64 |

*Table 3: Values of some evaluation's criteria*

Figure 6 shows the evolution of MAE and RMSE loss values for the different architectures, with *timestep=5* using (64×64) and (128×128) images respectively. It is observed that loss values in ConvLSTM model remain fairly constant after few epochs, this means that parameters have to be more optimized to improve performance model (more complexity). We can also note that the evolution of training loss for Stack-LSTM and CNN-LSTM models are almost similar using (64×64) images. However, when the sizes of images change to (128×128), there is a significant difference for the CNN-LSTM model which has the lowest values.

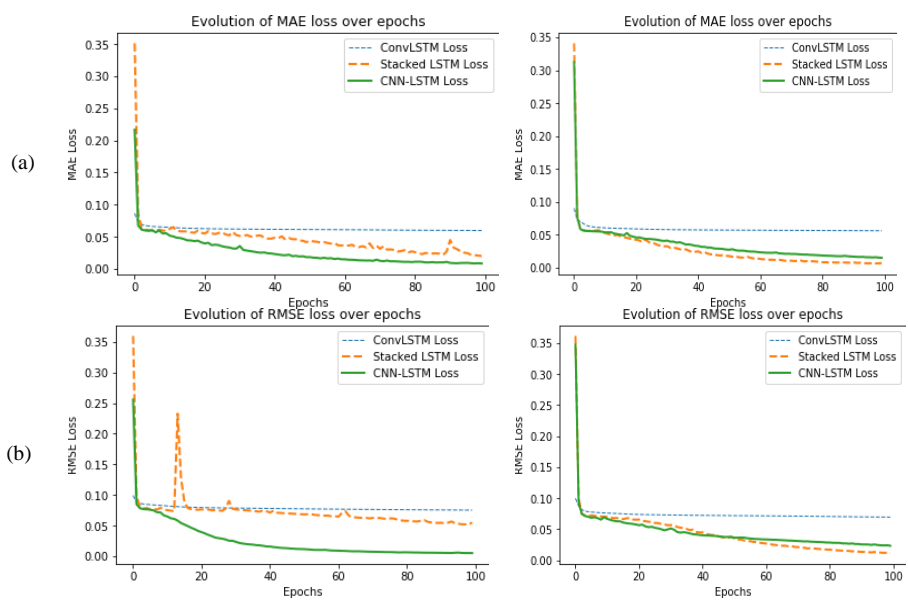

**Figure 6:** *Evolution of training loss values over epochs. (a) Left: MAE with (128×128) images, Right: MAE with (64×64) images. (b) Left: RMSE with (128×128) images, Right: RMSE with (64×64) images.*

## 5. Conclusion

In this study we made experimentation and tested several parameters in order to determine which LSTM architecture is suitable for the problem of prediction in remote sensing images time series. After the analysis of results, it is noted that the use of ConvLSTM architecture for this kind of problem is not advisable. When size of images and length of sequences become higher, this architecture does not perform well and results are not very satisfactory compared to Stack-LSTM and CNN-LSTM. To expect good results with ConvLSTM architecture, high parameters values should be chosen and therefore, much memory resources. In addition to that, because of convolutions operations, time processing is significantly higher than with CNN-LSTM and Stack-LSTM. However, although processing time and training loss are the lowest with stack-LSTM architecture in some cases, CNN-LSTM seems to produce better results by analyzing all others parameters. In fact, in LSTM model, predictions are made pixel by pixel while in CNN-LSTM, the CNN part of model extract important features and then the LSTM network memorize how they are changing over the time. Thus, use of CNN-LSTM architecture is recommended for forecasting tasks using earth observation images time series. Nevertheless, just knowing which architecture use for solving a problem is not enough. In all situations it is necessary to choose better parameters to achieve good results. And thus, next challenge for our study is to determine how to optimize model to reach as much as possible the best accuracy.

[6] Changjiang Xiao, Nengcheng Chen, Chuli Hu, Ke Wang, Zewei Xu, Yaping Cai, LeiXu, Zeqiang Chen, and Jianya Gong. "A spatiotemporal deep learning model for sea surface temperature field prediction using time-series satellite data". *Environmental Modelling & Software*, 120:104502, 2019.

[7] Hassan Ismail Fawaz, Germain Forestier, Jonathan Weber, Lhassane Idoumghar, and Pierre-Alain Muller. "Deep learning for time series classification: a review". *Data Mining and Knowledge Discovery*, 33(4):917–963, 2019.

[8] Lardeux Cédric, Kemavo Anoumou, Rageade Maxence, Rham Mathieu, Frison Pierre-Louis, and Rudant Jean-paul. "Mise en œuvre d'outils open source pour le suivi opérationnel de l'occupation des sols et de la déforestation à partir des données Sentinel radar et optique". *Revue Française de Photogrammétrie et Télédétection*, volume=220, pages=59–70, 2019.

[9] Qi Yang, Liangsheng Shi, Jinye Han, Yuanyuan Zha, and Penghui Zhu. "Deep convolutional neural networks for rice grain yield estimation at the ripening stage using uav-based remotely sensed images". *Field Crops Research*, 235:142–153, 2019.

[10] S. Hong, S. Kim, M. Joh, and S.-k. Song, "Psique: Next sequence prediction of satellite images using a convolutional sequence-to sequence network," arXiv preprint arXiv:1711.10644, 2017

[11] S. Kim, S. Hong, M. Joh, and S.-k. Song, "Deeprain: Convlstm network for precipitation prediction using multichannel radar data," arXiv preprint arXiv: 1711.02316, 2017.

[12] Sepp Hochreiter and Jürgen Schmidhuber. "Long short-term memory". *Neural computation*, 9(8):1735–1780, 1997.

[13] SHI Xingjian, Zhourong Chen, Hao Wang, Dit-Yan Yeung, Wai-Kin Wong, and Wang-chun Woo. "Convolutional lstm network: A machine learning approach for precipitation nowcasting". *In Advances in neural information processing systems*, pages 802–810, 2015.

[14] Xing Fang and Zhuoning Yuan. "Performance enhancing techniques for deep learning models in time series forecasting". *Engineering Applications of Artificial Intelligence*, 85:533–542, 2019.

[15] Z. Shen, Y. Zhang, J. Lu, J. Xu, and G. Xiao, "A novel time series forecasting model with deep learning," Neurocomputing, 2019.

[16] Zhou Wang, Alan C Bovik, Hamid R Sheikh, Eero P Simoncelli, et al. "Image quality assessment: from error visibility to structural similarity". *IEEE transactions on image processing*, 13(4):600–612, 2004.